\documentclass{article}

\usepackage{microtype}
\usepackage{graphicx}
\usepackage{subcaption}
\usepackage{booktabs} 

\usepackage{hyperref}

\usepackage[accepted]{icml2026}

\usepackage{amsmath}
\usepackage{amssymb}
\usepackage{mathtools}
\usepackage{amsthm}

\usepackage[capitalize,noabbrev]{cleveref}

\theoremstyle{plain}

\theoremstyle{definition}

\theoremstyle{remark}

\usepackage[textsize=tiny]{todonotes}

\icmltitlerunning{Mahalanobis-Guided Latent OOD Detection for Hybrid ES--DRL Control in Time-Varying Systems}

\begin{document}

\twocolumn[

 \icmltitle{Mahalanobis-Guided Latent OOD Detection for Hybrid ES--DRL Control in Time-Varying Systems}



  \icmlsetsymbol{equal}{*}

  \begin{icmlauthorlist}
    \icmlauthor{Shaifalee Saxena}{equal,lab,sch}
    \icmlauthor{Alexander Scheinker}{equal,lab}
  \end{icmlauthorlist}

  \icmlaffiliation{lab}{Los Alamos National Laboratory, Los Alamos, New Mexico, USA}
  \icmlaffiliation{sch}{University of New Mexico, New Mexico, USA}

  \icmlcorrespondingauthor{Shaifalee Saxena}{shaifalees@lanl.gov}
  \icmlcorrespondingauthor{Alexander Scheinker}{ascheink@lanl.gov}

  \icmlkeywords{Machine Learning, ICML}

  \vskip 0.3in
]



\printAffiliationsAndNotice{}  

\begin{abstract}
In this paper, we study Mahalanobis-guided latent out-of-distribution (OOD) detection for test-time RL controller switching in nonlinear time-varying systems. RL controllers can quickly control high-dimensional systems within the training distribution, but their performance can degrade when time-varying dynamics produce unseen observations. We consider a combined ES--DRL controller, where RL provides fast in-distribution actions and bounded extremum seeking (ES) provides robust model-independent control under OOD operation. The key challenge is deciding when to switch. We train a variational autoencoder (VAE) on in-distribution beam-profile observations and use Mahalanobis distance in the VAE latent space to detect OOD beam profiles at test time. This OOD decision sets a binary switch that selects either the RL controller or the ES controller. We evaluate the approach in safety-critical particle accelerator control. In this setting, spatial magnet motion creates OOD beam profiles that were not seen during RL training. Visualization of the VAE latent space shows that the proposed method identifies this OOD scenario and provides an interpretable signal for switching between RL and ES in the combined controller.
\end{abstract}

\section{Introduction}
Reinforcement learning (RL) is a promising approach for high-dimensional control because it can learn fast policies through interaction in simulation \cite{sutton2018reinforcement}, and has seen great success in real-world robotic systems \cite{gu2017deep}. However, learned RL policies are typically reliable only near the distribution on which they were trained ~\citep{danesh2021oodd}. When the system evolves over time because of actuator drift, calibration error, changing operating conditions, or geometric perturbations, the deployed observations may become out-of-distribution (OOD), leading to degraded or unsafe actions.

Many ongoing efforts exist to improve the safety and robustness of deep learning methods in general and RL methods in particular for changing systems. Action-constrained RL (ACRL) is a generic framework for learning control policies with zero action constraint violation, which is required by various safety-critical and resource-constrained application \cite{hung2025efficient}. Safe reinforcement learning is an approach to RL for real-world problems in which unsafe states can be avoided by planning ahead a short time into the future when a sufficiently accurate model can avoid unsafe states \cite{thomas2021safe}. The difficulty of designing deep RL algorithms for novel problems is being studied with new automated RL frameworks \cite{parker2022automated}. For difficulties faced by LLMs in adapting to new data distributions retrieval-augmented LMs are being studied \cite{asai2024reliable}. 

Nonlinear time-varying systems make this problem particularly important because deployment conditions may differ from those seen during training. A hybrid controller is useful in this setting because it combines the complementary strengths of learning-based and model-free control. The RL policy can produce fast, coordinated actions for high-dimensional systems when the observation remains close to the training distribution. In contrast, a robust model-free controller such as extremum seeking (ES) can continue to adapt when the dynamics drift, but may converge more slowly in large action spaces. The hybrid architecture therefore uses RL for rapid in-distribution control and ES for robustness when the system becomes uncertain or distributionally unfamiliar.

Particle accelerator control is a representative example of this challenge. Particle accelerators support a wide range of scientific applications, including neutron production, materials characterization, nuclear physics, medical isotope production, and high-energy physics experiments. These systems are safety-critical: poor control actions can increase beam loss, drive the beam toward aperture limits, or degrade experimental operation. As a result, learned controllers must be deployed carefully and should not be trusted blindly when beam dynamics move outside the conditions seen during training. In our setting, the RL controller acts on quadrupole magnet strengths, while the supervisor monitors beam-profile observations. Time-varying magnet motion and geometry changes can push the beam dynamics outside the RL training distribution.

The central challenge is selecting the switch between RL and ES at test time. We investigate Mahalanobis-guided latent OOD detection for RL controller switching. A probabilistic latent model is trained on in-distribution beam-profile observations and used to embed each test-time observation into a low-dimensional latent space. The Mahalanobis distance between the current latent representation and the in-distribution latent model is then used to select the binary switching coefficient $\beta_t$. Small distances select the RL controller, while large distances indicate OOD behavior and trigger a switch to ES. In this way, the supervisor in the combined ES--DRL controller is formulated as a latent OOD detection problem.

\section{Related Work}

\paragraph{Hybrid RL and fallback control.}
Hybrid control architectures combine the fast decision-making capability of learned RL policies with the robustness of classical, adaptive, or model-free controllers. Recent work combines DRL with bounded extremum seeking to improve robustness in nonlinear time-varying systems, including accelerator tuning, where RL provides fast nominal control while the fallback controller maintains performance under drift~\citep{saxena2025improved}. A related hybrid controller has also been studied for robotic manipulation under distribution shift, including time-varying goals and spatially varying friction~\citep{saxena2026deep}. Other hybrid RL-control approaches combine learned policies with model predictive control or adaptive control, for example through actor-critic MPC~\citep{romero2025actor} or MRAC-RL~\citep{guha2021mracrl}. These methods motivate hybrid control as a practical architecture for deploying RL in changing environments. In contrast, our focus is not on designing a new fallback controller, but on learning the switching supervisor that decides when the RL policy should be trusted.


\paragraph{OOD and anomaly detection in reinforcement learning.}
Detecting OOD observations is important for reliable RL deployment because unfamiliar states can cause learned policies to select low-performance or unsafe actions. Early work formulated OOD classification in deep RL using uncertainty estimates and policy entropy~\citep{sedlmeier2020uncertainty,sedlmeier2020policyentropy}. Other work studies OOD dynamics detection, where the objective is to identify changes in the environment dynamics relative to the training distribution~\citep{danesh2021oodd}. Probabilistic dynamics models and bootstrapped ensembles have also been used to detect OOD situations for RL agents~\citep{haider2023ood}. More recently, OOD detection in RL has been revisited with benchmarks that include temporally correlated anomalies and time-series-based detection methods~\citep{nasvytis2024rethinking}. These works focus on detecting anomalous states or dynamics. Our work uses OOD detection as an actionable control signal: the latent distance determines the binary switching coefficient $\beta_t$ in a hybrid controller.
\begin{figure*}[]
    \centering
    \includegraphics[width=0.8\linewidth]{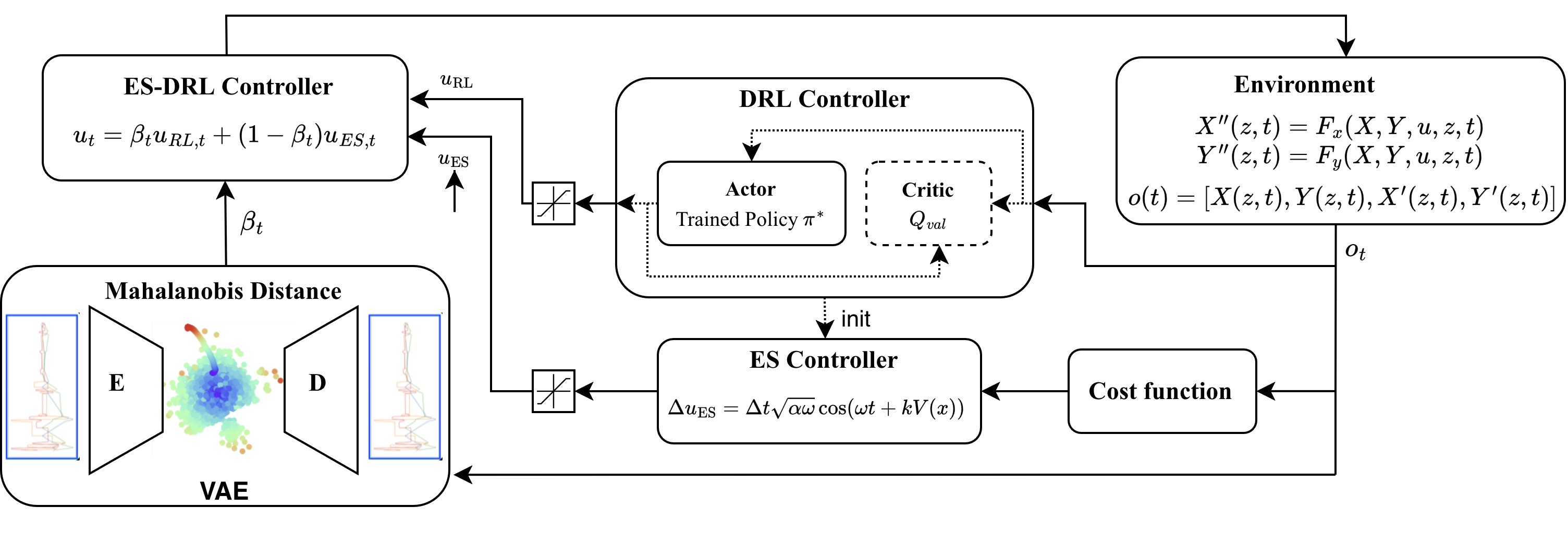}
    \caption{VAE-guided ES/RL switching setup.}
    \label{fig:ESRL}
\end{figure*}

\begin{figure*}[]
    \centering
    \includegraphics[width=0.9\linewidth]{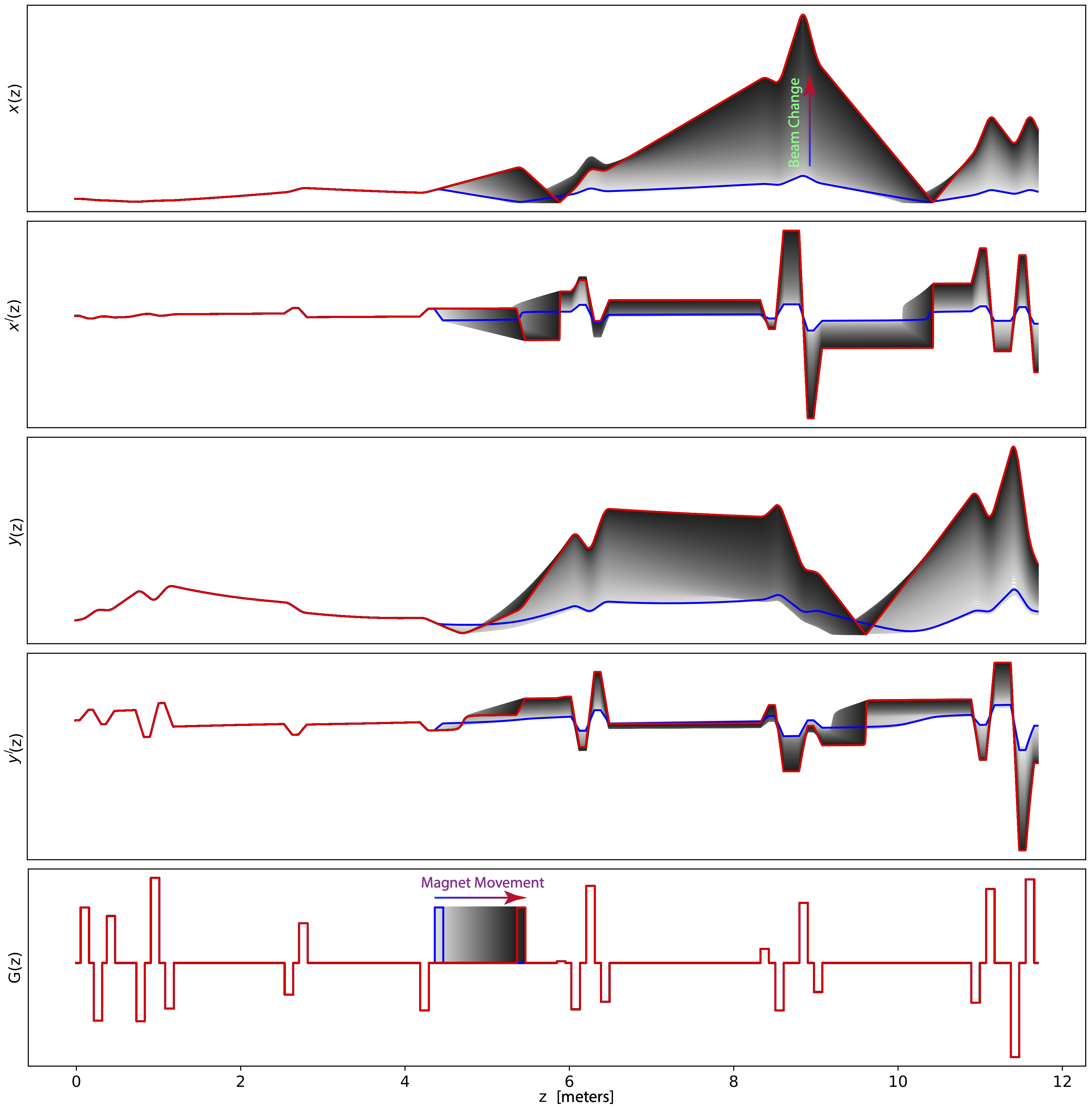}
    \caption{Effect of spatial magnet movement on beam profiles. The moving magnet creates non-smooth envelope and slope behavior with large excursions. The $z$ coordinate is in meters; other values are normalized.}
    \label{fig:MagMove}
\end{figure*}

\paragraph{Mahalanobis and latent-space anomaly detection.}
Mahalanobis distance is widely used to measure whether a feature vector lies far from a reference distribution. In supervised deep learning, class-conditional Gaussian feature models have been used to detect OOD and adversarial samples with a Mahalanobis confidence score~\citep{lee2018mahalanobis}. In reinforcement learning, MDX extends Mahalanobis-distance detection to RL by estimating class-conditional feature distributions from policy-network representations and detecting random, adversarial, and OOD state outliers~\citep{zhang2024distance}. Our approach builds on this distance-based perspective but differs in both representation and use. We compute Mahalanobis distance in a learned latent representation of physical system observations, rather than in action-class features of a policy network, and use the resulting distance to select between the RL and ES controllers.

\section{Problem Formulation}

\subsection{Particle Accelerator Tuning Problem}

Particle accelerator tuning is challenging because the beam dynamics are nonlinear, strongly coupled, and sensitive to both magnet settings and incoming beam conditions. A change in one quadrupole magnet can affect the downstream beam envelope, and poor settings can produce large envelope excursions, non-smooth beam evolution, and beam loss. The controller must therefore tune many coupled actuators while maintaining a compact, smooth, and well-aligned beam.

We evaluate the proposed supervisor in a particle accelerator tuning problem based on the Kapchinskij--Vladimirskij (KV) envelope model~\citep{kapchinskij1959limitations}. The system represents a low-energy beam transport section of a linear accelerator. The beam state is described by the horizontal and vertical envelope radii $X(z,t)$ and $Y(z,t)$ and their slopes $X'(z,t)$ and $Y'(z,t)$ along the beamline coordinate $z$. The controller acts on quadrupole magnet strengths.

At each control step, the observation is the sampled beam-profile vector
\(
o_t =
\left[
X(z,t),\;Y(z,t),\;X'(z,t),\;Y'(z,t)
\right],
\)
and the control input is the vector of setpoints for the 22 quadrupole magnet strengths,
\[
Q_t =
[Q_1(t),\ldots,Q_{22}(t)]^\top
\in \mathbb{R}^{22}.
\]
The RL controller is trained using Deep Deterministic Policy Gradient (DDPG) \cite{lillicrap2020continuous} in simulation over a finite set of beamline configurations and operating regimes. During training, beam initial conditions are randomized across episodes so that the policy observes different initial beam profiles. The actor is trained to maximize a reward that encourages compact beam envelopes, smooth beam evolution, and terminal alignment. During test time, the trained actor is frozen to prevent policy drift. Additional DDPG training details and the reward definition are provided in Appendix~\ref{ddpg}.

\subsection{OOD-Aware Test-Time RL--ES Switching}

We combine the trained RL actor with bounded extremum seeking (ES) because the two controllers have complementary strengths. The RL actor provides fast, coordinated magnet commands when the observed beam profile is close to the distribution encountered during DDPG training. ES provides robustness under unknown and time-varying dynamics because it optimizes a measured scalar objective without requiring an analytic model of the accelerator, while keeping parameter updates bounded
\citep{scheinker2013model,scheinker2016bounded}. It has also been demonstrated for particle accelerator beam-loss minimization \citep{scheinker2021extremum}. However, ES is a local feedback-based optimizer: in high-dimensional tuning problems, such as the 22-dimensional quadrupole setting considered here, convergence can be slow and the search may settle in suboptimal local regions. Thus, using ES alone can be unnecessarily slow, while using RL alone can be unreliable under OOD conditions.

We use RL in two ways. First, when the beam profile is detected as in-distribution, the RL actor directly supplies the control action. Second, the ES controller is warm-started from the RL-recommended magnet setting, reducing the transient associated with starting local search from an arbitrary magnet setting. ES uses the same beam-quality objective as the RL reward and produces an action $u_{\mathrm{ES},t}$ for the quadrupole magnet strengths. 

The hybrid controller selects between the RL and ES actions through a binary switch,
\begin{equation}
    u_t = \beta_t u_{\mathrm{RL},t} + (1-\beta_t)u_{\mathrm{ES},t}, \qquad \beta_t \in \{0,1\}.
    \label{eq:esrl}
\end{equation}

When $\beta_t=1$, the RL action is applied. When $\beta_t=0$, the ES action is applied. Thus, $\beta_t$ is not a continuous authority weight in this work; it is a test-time selector between the two controllers.

 We select $\beta_t$ using latent OOD evidence. As shown in Fig.~\ref{fig:ESRL}, during test-time deployment the supervisor observes the current beam profile, embeds it into the learned latent space, computes its Mahalanobis distance from the training-distribution latent model, and uses this score to select either the RL controller or the ES controller. 
 
\section{Mahalanobis-Guided Latent Supervisor}

\begin{figure*}[]
    \centering
    \includegraphics[width=1.0\linewidth]{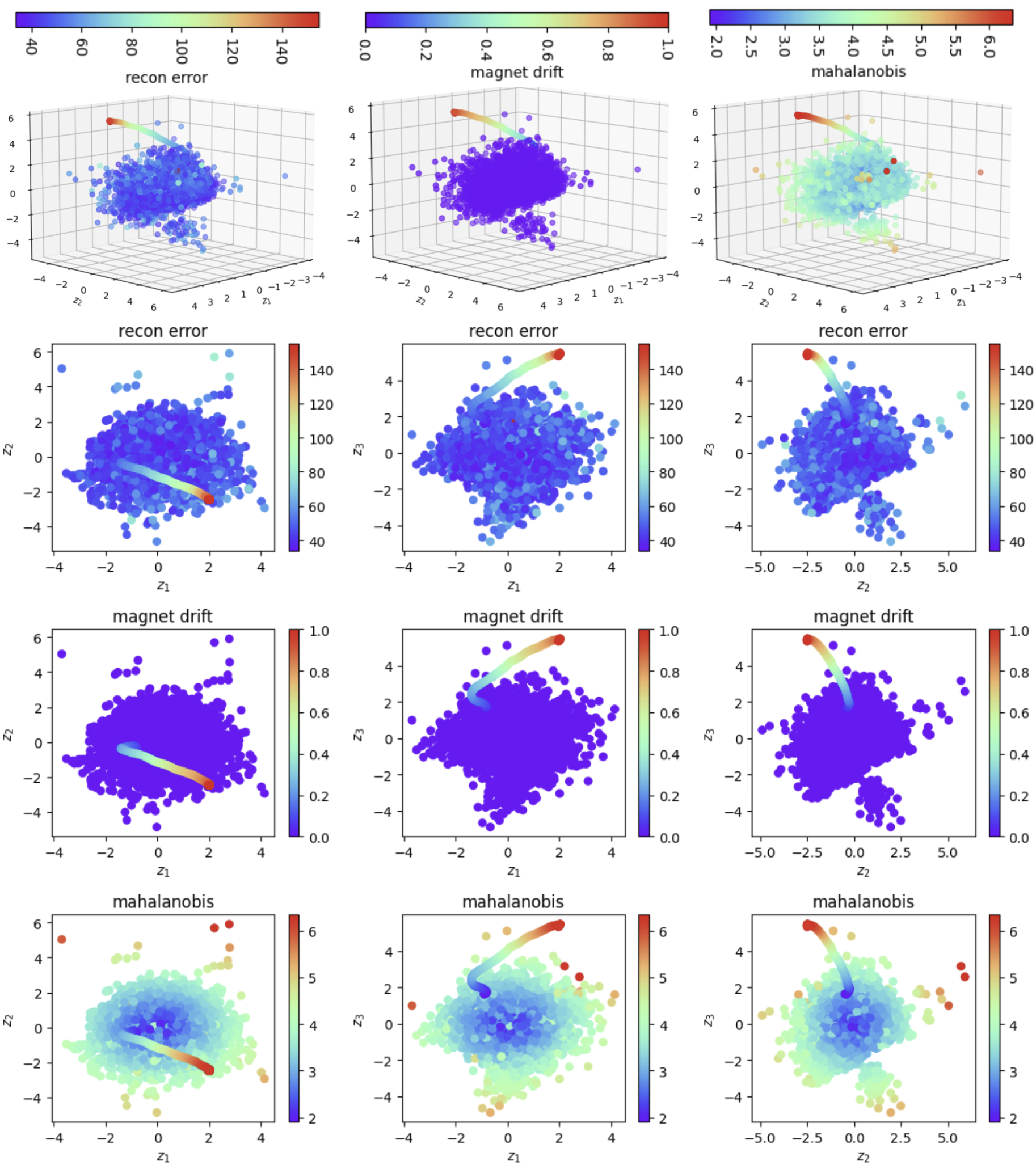}
    \caption{The top row shows the locations of 8192 embedded test beam envelopes in the 3D latent space of the trained VAE together with the embeddings of 128 beam envelopes from a time-varying lattice in which one of the magnets moves 1 meter. The points are colored by reconstruction error, by how far the magnet has moved, and by the Mahalanobis distance of each point from the latent distribution learned by the VAE which is modeled as $\mathcal{N}(\mathbf{0},I_{3\times 3})$. The next three rows show orthogonal 2D projections of the 3D view.}
    \label{fig:Ldim3}
\end{figure*}

\subsection{Learning a Latent Model of Beam Profiles}

The VAE-based switching supervisor is trained to learn a compact representation of in-distribution beam-profile observations. To construct the VAE dataset, we start from an expert-tuned quadrupole setting that produces a stable beam profile. We then generate beam-profile data by applying random perturbations around this reference quadrupole setting and solving the KV envelope model. This produces a dataset of $700{,}000$ beam profiles, which was split into $680{,}000$ training, $10{,}000$ validation, and $10{,}000$ test profiles. VAE architecture, training details, and latent-dimension comparison are provided in Appendix~\ref{vae_arch}.

The OOD detector takes the beam-profile observation as input. Since the beam profile is sampled along the longitudinal coordinate $z$, the input has four channels corresponding to $X(z,t)$, $Y(z,t)$, $X'(z,t)$, and $Y'(z,t)$. Each channel is sampled at $N_z=4000$ longitudinal locations, so each input beam profile is represented as
\[
x_t \in \mathbb{R}^{4 \times 4000}.
\]
We train a variational autoencoder (VAE) \cite{kingma2013auto} to embed these high-dimensional beam profiles into a low-dimensional latent space. The encoder is implemented using one-dimensional convolutional layers along the beamline coordinate, reducing the spatial resolution as
\(
4000 \rightarrow 2000 \rightarrow 1000 \rightarrow 500
\rightarrow 250 \rightarrow 125 \rightarrow \mathrm{Dense}
\rightarrow L_{\mathrm{dim}}.
\)

The dense layers output the latent mean and variance,
\[
\mu_\phi(x_t),\;\sigma_\phi^2(x_t) \in \mathbb{R}^{L_{\mathrm{dim}}}.
\]
As shown in Figure~\ref{fig:Ldim3}, we use $L_{\mathrm{dim}}=3$.

The encoder maps each beam profile to a Gaussian latent posterior,
\[
q_\phi(z_t|x_t)
=
\mathcal{N}
\left(
\mu_\phi(x_t),
\operatorname{diag}(\sigma_\phi^2(x_t))
\right),
\]
and the decoder reconstructs the beam profile from the latent variable. The VAE is trained by minimizing
\begin{equation}
\mathcal{L}_{\mathrm{VAE}} =
\mathbb{E}_{q_\phi(z|x)}
\left[-\log p_\psi(x|z)\right]
+
\lambda_{\mathrm{KL}}
D_{\mathrm{KL}}
\left(q_\phi(z|x)\|p(z)\right),
\label{eq:vae}
\end{equation}
where $p(z)=\mathcal{N}(0,I)$.

\subsection{Latent Mahalanobis OOD Score}

After training the VAE, a beam profile $x_t$ is embedded using the encoder mean,
\[
z_t = \mu_\phi(x_t).
\]
The VAE prior regularizes the latent space toward a standard Gaussian distribution. In the reported experiments, we use a three-dimensional latent space and model the reference latent distribution as
\[
\mathcal{Z}_{\mathrm{ID}} = \mathcal{N}(0,I_3).
\]
More generally, the reference mean and covariance can be estimated from training-distribution embeddings,
\[
\mathcal{Z}_{\mathrm{ID}}
=
\mathcal{N}
\left(
\bar{z}_{\mathrm{ID}},
\Sigma_{\mathrm{ID}}
\right).
\]
For the standard Gaussian reference used in Fig.~\ref{fig:Ldim3}, this corresponds to $\bar{z}_{\mathrm{ID}}=0$ and $\Sigma_{\mathrm{ID}}=I_3$.

We compute the squared Mahalanobis distance
\begin{equation}
d_M^2(x_t)
=
(z_t-\bar{z}_{\mathrm{ID}})^\top
\left(\Sigma_{\mathrm{ID}}+\epsilon I\right)^{-1}
(z_t-\bar{z}_{\mathrm{ID}}),
\label{eq:dM}
\end{equation}
where $\epsilon I$ is a small regularization term for numerical stability. This score is small when the current beam profile lies near the training-distribution latent cloud and large when the beam profile is distributionally unfamiliar.

The Mahalanobis score is used as the OOD evidence for test-time switching. As shown in Fig.~\ref{fig:Ldim3}, time-varying magnet drift produces increasing Mahalanobis distance in latent space.

The reconstruction error used to color the ``recon error" plots in Fig.~\ref{fig:Ldim3} is a simple absolute value sum over channels and accelerator locations, calculated as:
\begin{equation} 
E = \sum_{c=1:4}\sum_{z=1:4000} |\hat{o}(z,c) - o(z,c)|.\label{recon_error}
\end{equation}

\subsection{Binary Test-Time Switch}

The Mahalanobis score is used to set the binary switch $\beta_t$. We use
\begin{equation}
\beta_t =
\begin{cases}
1, & d_M^2(x_t) \leq \tau,\\
0, & d_M^2(x_t) > \tau,
\end{cases}
\label{eq:beta}
\end{equation}
where $\tau$ is the OOD threshold chosen from the training-distribution latent distances.

Thus, $\beta_t=1$ selects the RL controller when the beam profile is close to the training-distribution latent model, while $\beta_t=0$ selects ES when the beam profile is detected as OOD.

The applied control is given by Eq.~\ref{eq:esrl}. The VAE acts as a test-time supervisor that determines whether the current beam profile is reliable for the trained RL policy or whether control should switch to ES.

\section{Results}

\paragraph{Physical effect of magnet drift.}
During training, the quadrupole magnet locations are fixed. At test time, one quadrupole magnet is moved spatially along the beamline, creating a time-varying lattice that changes the beam dynamics. This perturbation is therefore an OOD condition for the trained RL controller: although the RL actor has learned to tune magnet strengths for the training lattice, it has not seen beam profiles generated by a moving magnet.

Figure~\ref{fig:MagMove} illustrates the physical effect of this perturbation. As the magnet is moved, the beam profiles transition from smooth, compact envelopes to non-smooth profiles with large excursions. In particular, the horizontal and vertical envelopes and their slopes develop sharp changes near and downstream of the moving magnet. These changes indicate that the beam is no longer following the smooth behavior encouraged by the RL reward. When the envelope grows toward or beyond the allowed aperture, the beam effectively hits the beam-pipe wall, which corresponds to high beam loss and low reward. In this regime, the RL action should not be trusted because the observed beam profile is outside the distribution encountered during training.

\paragraph{Latent OOD evidence.}
To quantify the distribution shift, we pass beam-profile observations through the trained VAE and analyze their embeddings in the three-dimensional latent space. Figure~\ref{fig:Ldim3} shows 8192 embedded test beam envelopes from the training-distribution lattice together with 128 beam envelopes generated during the time-varying magnet-drift trajectory.

The training-distribution beam profiles form a compact latent cloud. In contrast, the moving-magnet trajectory appears as a structured path that leaves this cloud as the magnet drift increases. This is important because the VAE is not using the magnet position directly as an input; it only observes the resulting beam profile. Therefore, the separation in latent space indicates that the beam-profile measurements themselves contain enough information to reveal the OOD condition.

Figure~\ref{fig:Ldim3} also shows three complementary OOD indicators. First, the reconstruction error increases along the drift trajectory, indicating that the VAE has more difficulty reconstructing beam profiles produced by the moving lattice. Second, the magnet-drift coloring confirms that the trajectory moves progressively as the magnet is displaced from its training-time location. Third, the Mahalanobis distance increases as the trajectory moves away from the training-distribution latent cloud. In the plotted color scale, the central training-distribution region is mostly associated with lower Mahalanobis values, roughly in the range $2$--$4$, while the high-drift points move toward larger values, approximately $5$--$6$.

These trends connect the physical and latent views of OOD behavior. In Figure~\ref{fig:MagMove}, increasing magnet movement produces non-smooth beam envelopes and potential beam loss. In Figure~\ref{fig:Ldim3}, the same perturbation produces increasing reconstruction error and increasing Mahalanobis distance. Thus, the Mahalanobis distance in the VAE latent space provides a compact test-time signal that reflects physically meaningful degradation in the beam profile.

\paragraph{Implication for switching.}
The Mahalanobis score is used to choose the binary switch $\beta_t$. For the three-dimensional latent model in Figure~\ref{fig:Ldim3}, the plotted values show separation between the dense training-distribution latent cloud and the high-drift portion of the moving-magnet trajectory. This suggests that an OOD threshold $\tau$ can be selected to distinguish beam profiles that remain close to the training-distribution behavior from those that move into the higher-distance OOD region.

With this thresholding rule, beam profiles close to the training-distribution latent cloud select the RL controller, while profiles with larger Mahalanobis distance select ES. The threshold $\tau$ should be interpreted as a calibration parameter rather than a universal constant. A lower threshold would switch to ES earlier, while a higher threshold would keep RL active longer but may delay intervention.

Overall, Figures~\ref{fig:MagMove} and~\ref{fig:Ldim3} show that spatial magnet drift creates both a physical signature of OOD behavior and a corresponding latent-space signature. The physical signature is the transition to non-smooth beam envelopes and increased beam-loss risk. The latent signature is the increase in reconstruction error and Mahalanobis distance. This supports the use of latent Mahalanobis distance as the test-time switching signal between RL and ES.

\section{Conclusion}

We presented a Mahalanobis-guided latent OOD supervisor for test-time switching in a combined ES--DRL controller for nonlinear time-varying systems. A VAE is trained to learn a compact latent representation of in-distribution beam-profile observations, and the Mahalanobis distance in this latent space is used to decide whether the trained RL policy should be trusted or whether control should switch to ES. This allows the supervisor to use the full beam-profile structure rather than relying only on hand-designed physical thresholds. In the accelerator tuning problem, spatial magnet motion creates beam profiles that were not seen during RL training. These profiles become non-smooth, indicate increased beam-loss risk, and appear in the VAE latent space as a structured trajectory with increasing reconstruction error and Mahalanobis distance. These results show that the proposed supervisor can identify OOD beam-profile behavior and provide an interpretable switching signal between RL and ES. This work is a step toward safer deployment of RL controllers in time-varying physical systems. Future work will evaluate the complete closed-loop controller under broader accelerator perturbations and extend the framework so that detected OOD conditions can also trigger targeted RL fine-tuning, allowing the policy to adapt to newly observed operating regimes.

\nocite{langley00}

\bibliography{example_paper}
\bibliographystyle{icml2026}

\newpage
\appendix
\onecolumn
\section{DDPG training details}
\label{ddpg}
\begin{table}[h]
\centering
\caption{DDPG training settings for the accelerator controller.}
\label{tab:ddpg}
\begin{tabular}{ll}
\toprule
Quantity & Value \\
\midrule
Observation dimension & $4 \times 4000$ \\
Action dimension & $22$ \\
RL algorithm & DDPG \\
Discount factor & $0.99$ \\
Actor learning rate & $1\times 10^{-5}$ \\
Critic learning rate & $1\times 10^{-4}$ \\
Replay buffer size & $10^6$ \\
Batch size & $128$ \\
Target update & $\tau=0.005$ \\
Exploration noise & $\mathcal{N}(0,0.1)$ during training \\
Evaluation policy & Frozen actor, no exploration noise \\
\bottomrule
\end{tabular}
\end{table}

\paragraph{Reward function.}
The DDPG actor is trained with a beam-quality reward that encourages the beam to remain compact along the transport line, vary smoothly, and reach a well-aligned terminal profile. Let
\[
\langle a\rangle_+ = \max(0,a)
\]
denote the hinge function. For beam envelopes sampled at $N_z=4000$ longitudinal grid points, define the path-averaged beam sizes
\[
\bar{X}(t)=\frac{1}{N_z}\sum_{k=1}^{N_z}X(z_k,t),
\qquad
\bar{Y}(t)=\frac{1}{N_z}\sum_{k=1}^{N_z}Y(z_k,t).
\]
We set the envelope band and terminal target as
\[
r_{\mathrm{band}}=\frac{1}{2}r_{\max},
\qquad
r_{\mathrm{tt}}^2=\frac{1}{2}r_{\max}^2,
\]
where $r_{\max}$ is the operational beam-pipe radius.

The total penalty is
\[
P_t = P_{\mathrm{env}} + P_{\mathrm{smooth}} + P_{\mathrm{term}},
\]
with
\[
P_{\mathrm{env}}
=
w_e
\left(
\langle \bar{X}(t)-r_{\mathrm{band}}\rangle_+
+
\langle \bar{Y}(t)-r_{\mathrm{band}}\rangle_+
\right),
\]
\[
P_{\mathrm{smooth}}
=
w_s
\left(
\overline{X'^2}(t)+\overline{Y'^2}(t)
\right),
\]
where $\overline{X'^2}$ and $\overline{Y'^2}$ are path averages of the squared envelope slopes. The terminal penalty is
\[
P_{\mathrm{term}}
=
w_r
\left|X(z_{\max},t)-Y(z_{\max},t)\right|
+
w_w
\left(
\left|X'(z_{\max},t)\right|
+
\left|Y'(z_{\max},t)\right|
\right)
+
w_t
\left|
X^2(z_{\max},t)
+
Y^2(z_{\max},t)
-
r_{\mathrm{tt}}^2
\right|.
\]
The instantaneous reward is the bounded inverse
\[
R_t = \frac{1}{1+P_t} \in (0,1].
\]
Thus, the reward increases when the beam envelopes remain small, the slopes remain smooth, and the terminal beam profile is circular, flat, and close to the desired radius.

\newpage
\section{VAE Details}
\label{vae_arch}

Each input to the VAE has size $[4000,4]$ with 4000 locations along the beamline of $(x,x',y,y')$. A 1D Residual Convolutional Neural Network repeatedly decreases the input's initial size of 4000 by factors of 2 using 1D Convolutional layers with kernel size 3 and stride 2, while doubling the channel numbers. For the encoder, $[$tensor size, channel number$]$ are:
\[[4000,4] \rightarrow [4000,32] \rightarrow [2000,64] \rightarrow [1000,128] \rightarrow [500,256] \rightarrow [250,256] \rightarrow [125,256] \rightarrow [125, 1] \rightarrow [128] \rightarrow [L_d], \]
where the tensor of size $[125,1]$ was flattened and then passed through dense layers, transformed into a vector of size $[128]$, which was finally compressed to the latent dimension $L_d$. In our setup between each factor of 2 size reduction there are 2 residual blocks, with each block having 2 1D convolutional layers with the same number of channels as shown in that stage above, with each followed by a GroupNorm and a SiLU activation function. The mean and log-variance are then created from the last vector of size $[L_d]$ by one dense layer each before being passed to the sampling layer. The decoder architecture mirrors that of the encoder. The network was trained with a weight of 1e-3 on the KL divergence.

\begin{figure*}[]
    \centering
    \includegraphics[width=0.49\linewidth]
    {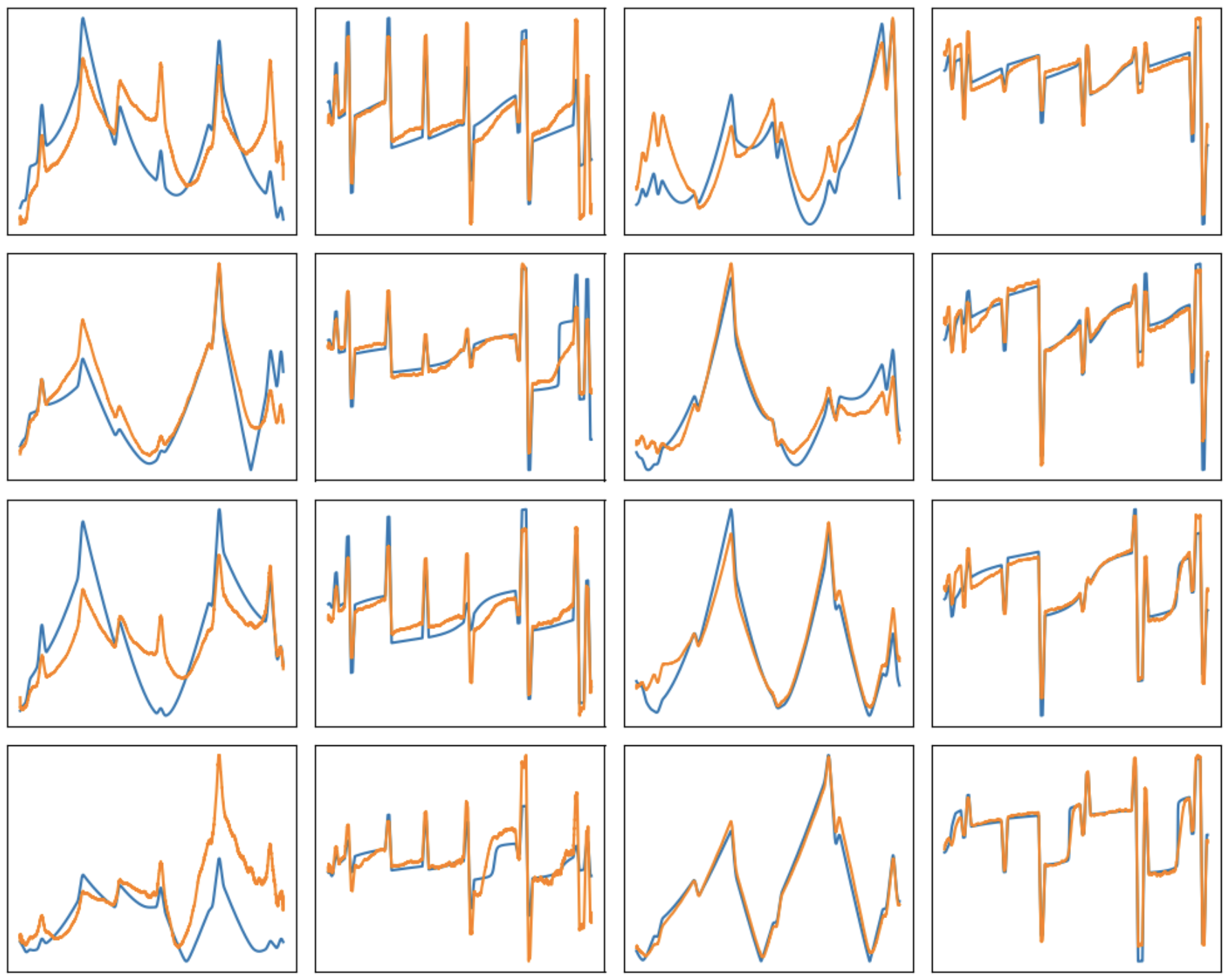}
    \includegraphics[width=0.49\linewidth]
    {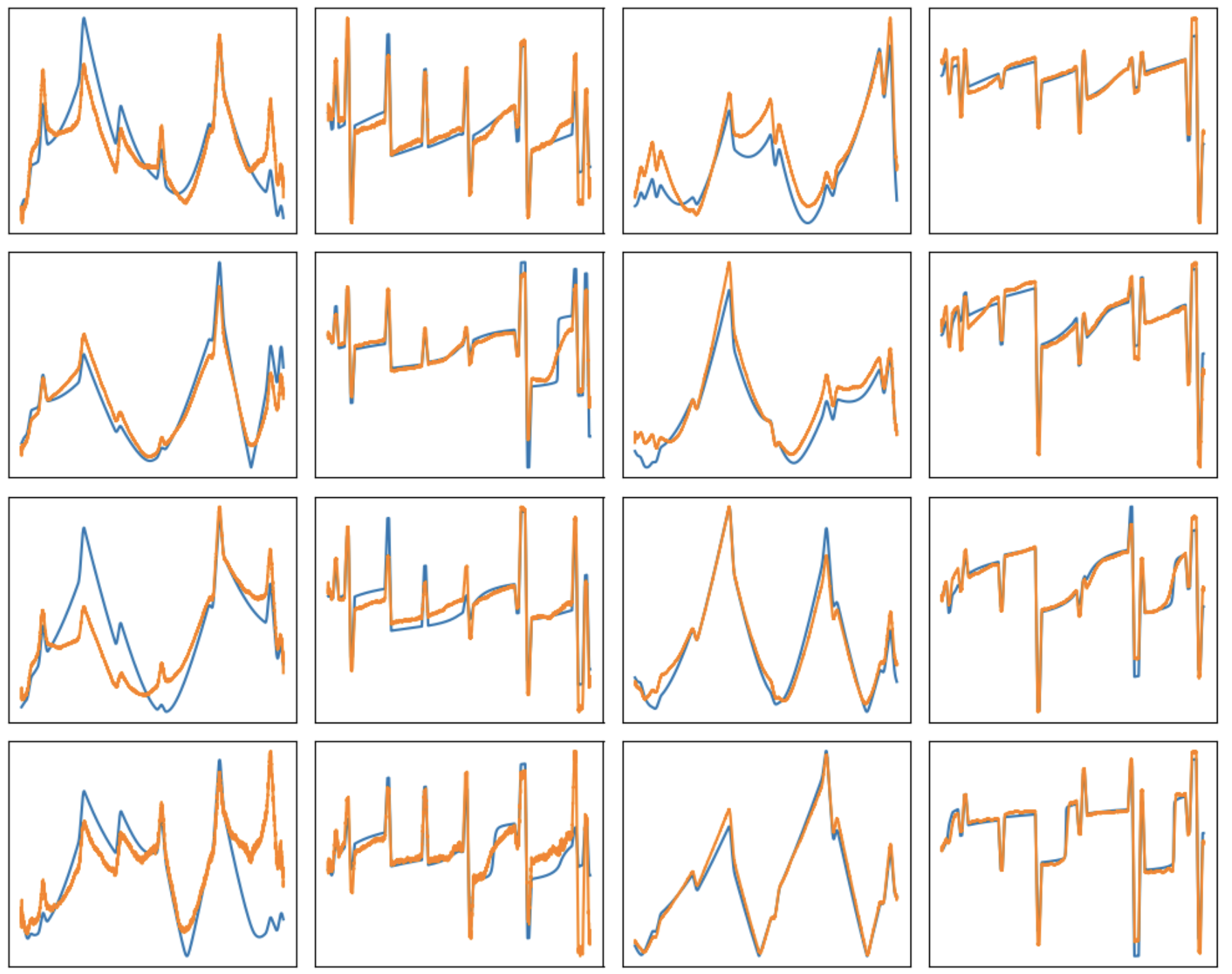}
    \caption{Left: The 4 rows show reconstructions of 4 random test samples shown on top of the correct values for the $L_d=2$ VAE. The 4 columns from left to right show the $x$, $x'$, $y$, and $y'$ beam envelopes. Right: The same is shown for $L_d=3$.}
    \label{fig:reconLdim2}
\end{figure*}

\begin{figure}[]
    \centering
    \includegraphics[width=0.42\linewidth]{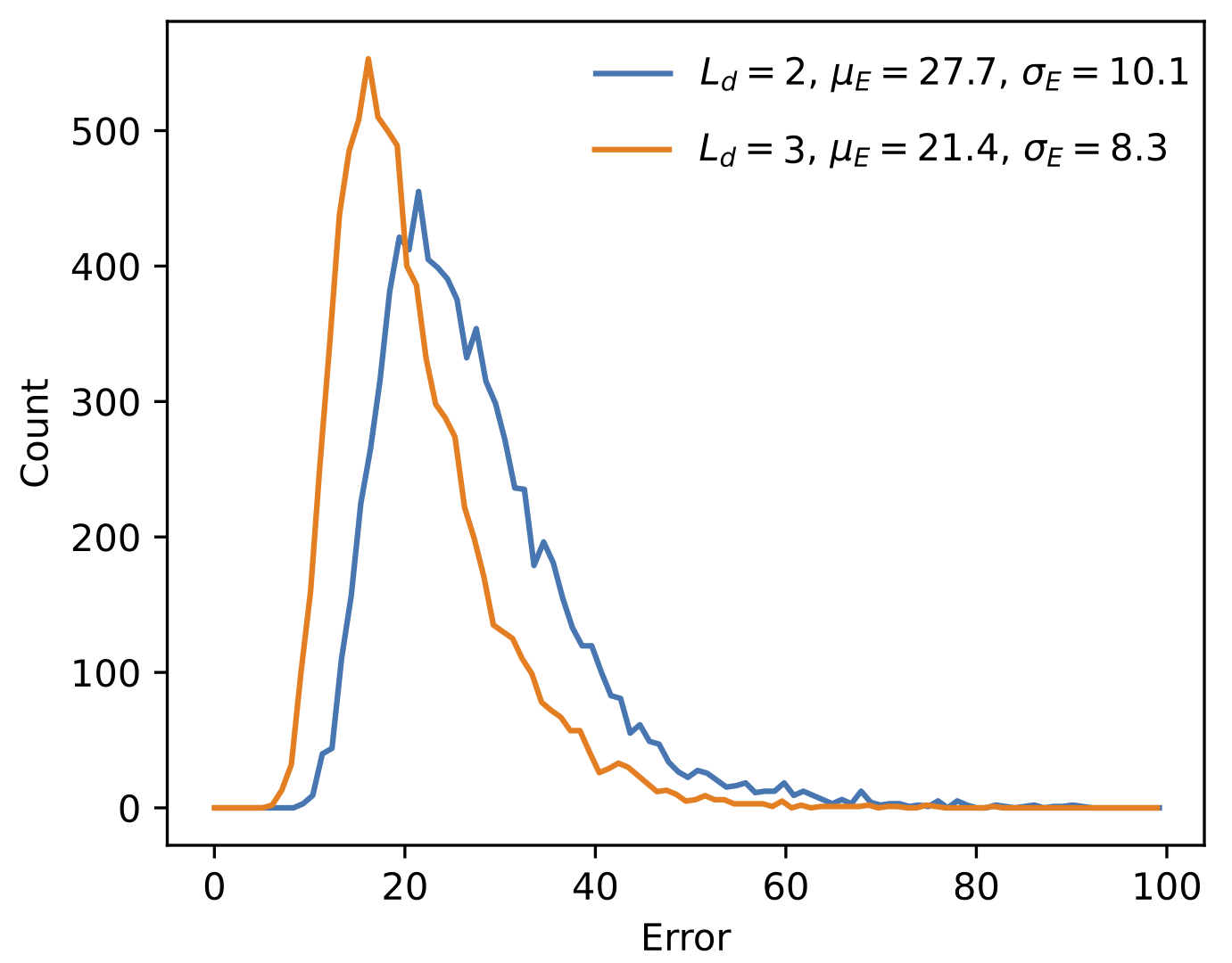}
    \includegraphics[width=0.55\linewidth]{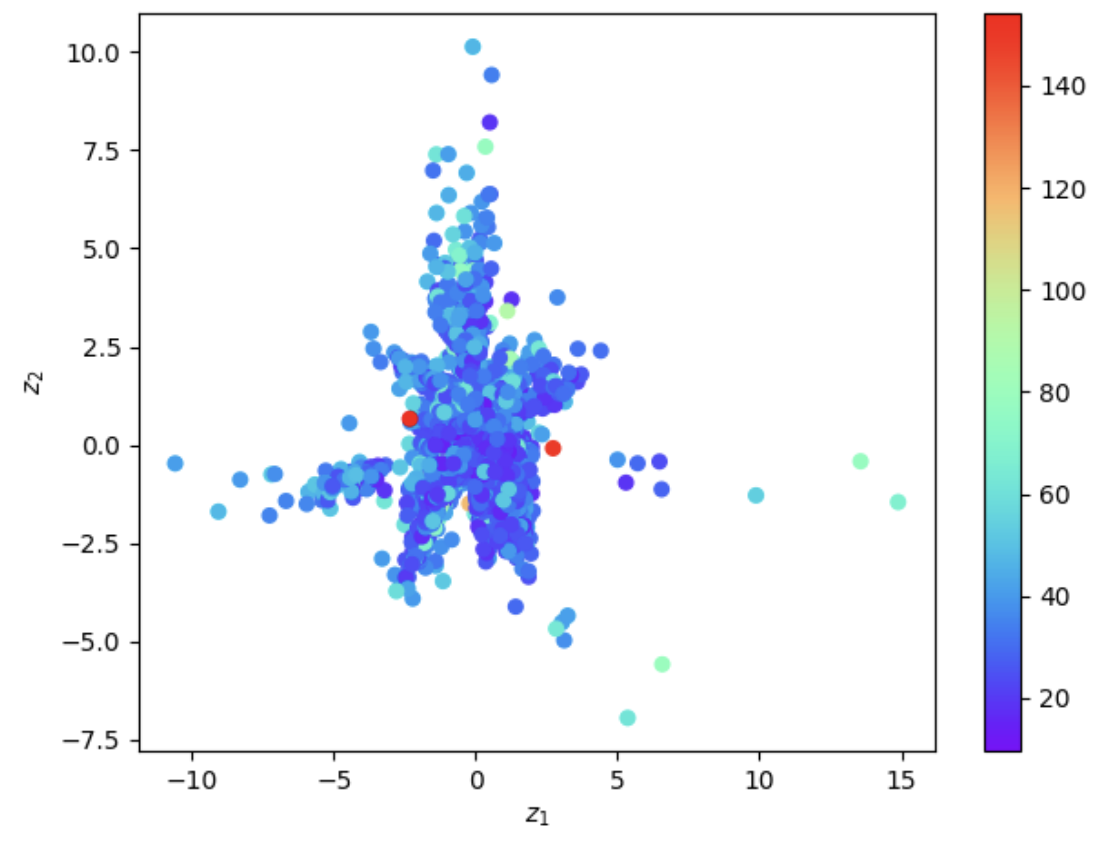}
    \caption{Left: Error statistics for 8192 test data points compared for the $L_d=2$ and $L_d=3$ VAEs. Right: Non-Gaussian latent space of the $L_d=2$ VAE colored by error.}
    \label{fig:hists}
\end{figure}

Our study found that with $L_d=2$ the model struggled to accurately reproduce the beam envelopes, as shown on the left side in Figure \ref{fig:reconLdim2} for 4 random samples from the test data. The VAE with $L_d=3$ was able to reproduce the beam envelopes more accurately, as shown on the right side in Figure \ref{fig:reconLdim2} for the same 4 samples. Furthermore, the higher accuracy of $L_d=3$ is quantified by the statistics of the reconstruction error from Equation \ref{recon_error} for 8192 test samples in Figure \ref{fig:hists}. Another difficulty with the $L_d=2$ model was a very non-Gaussian latent space as shown on the right side of \ref{fig:hists}, which would make Gaussian assuming Mahalanobis-based distance measurements inaccurate.

\end{document}